# The Minimum Information about CLinical Artificial Intelligence Checklist for Generative Modeling Research (MI-CLAIM-GEN)

Brenda Y. Miao[1*], Irene Y. Chen[2,3,4], Christopher YK Williams[1], Jaysón Davidson[1], Augusto Garcia-Agundez[5], Shenghuan Sun[1], Travis Zack[1,6], Suchi Saria[8,9,10,11], Rima Arnaout[1,2,12], Giorgio Quer[13], Hossein J. Sadaei[13,14], Ali Torkamani[13,14], Brett Beaulieu-Jones[15], Bin Yu[3,16,17], Milena Gianfrancesco[5], Atul J. Butte[1,7], Beau Norgeot[18, †], Madhumita Sushil[1, †]

1. Bakar Computational Health Sciences Institute, University of California, San Francisco, San Francisco, CA, USA
2. UCSF-UC Berkeley Joint Program in Computational Precision Health, University of California, Berkeley and University of California, San Francisco, Berkeley, CA, USA
3. Department of Electrical Engineering and Computer Sciences, University of California, Berkeley, Berkeley, CA, USA
4. Berkeley AI Research, University of California, Berkeley, Berkeley, CA, USA
5. Department of Medicine, Division of Rheumatology, University of California, San Francisco, San Francisco, California, USA
6. Helen Diller Family Comprehensive Cancer Center, University of California, San Francisco, San Francisco, CA, USA
7. Center for Data-driven Insights and Innovation, University of California, Office of the President, Oakland, CA
8. Bayesian Health, New York, NY 10282
9. Department of Computer Science, Johns Hopkins University Whiting School of Engineering, Baltimore, Maryland, USA
10. Department of Health Policy & Management, Johns Hopkins University Bloomberg School of Public Health, Baltimore, Maryland, USA
11. Department of Medicine, Johns Hopkins Medicine, Baltimore, MD 21205
12. Departments of Medicine, Radiology, and Pediatrics, University of California, San Francisco, San Francisco, CA, USA
13. Scripps Research Translational Institute, La Jolla, CA, USA
14. Department of Integrative Structural and Computational Biology, Scripps Research, La Jolla, CA 92037, USA.
15. Department of Medicine, University of Chicago, Chicago, IL, USA
16. Department of Statistics, University of California, Berkeley, Berkeley, CA, USA
17. Center for Computational Biology, University of California, Berkeley, Berkeley, CA, USA
18. Qualified Health PBC, Palo Alto CA

*Corresponding Author
Email: brenda.miao@ucsf.edu

Bakar Computational Health Sciences Institute, 490 Illinois Street
2nd Fl, North Tower, San Francisco, CA 94143

The "Minimum information about clinical artificial intelligence modeling" (MI-CLAIM) checklist[1], originally developed in 2020, provided a set of six steps with guidelines on the minimum information necessary to be reported about predictive artificial intelligence (AI) modeling studies to ensure transparent, reproducible research for AI in medicine.

Since then, recent advances in generative modeling, including large language models (LLMs), diffusion models, vision language models (VLMs), and other multimodal models have marked a paradigm shift in how machine learning models are being developed and deployed for biomedical research[2–4]. Many of these generative models are developed as foundation models and trained on large amounts of data and then augmented or further finetuned on smaller amounts of domain-specific data for specific clinical or biomedical tasks (Figure 1). The ability of these models to learn from limited amounts of domain-specific data, to integrate with external tools, and to perform complex tasks greatly exceed that of previous models. These changes necessitate the development of new guidelines for robust reporting of study design, model development, and evaluation, both pre and post-deployment, using clinical generative model research.

In response to gaps in standards and best practices for the reporting of clinical generative AI research identified by US Executive Order 14110[5] and several emerging national networks for clinical AI evaluation[6], we begin to formalize some of these guidelines by building on the original MI-CLAIM checklist. The new checklist, MI-CLAIM-GEN (Table 1), aims to address differences in training, evaluation, interpretability, and reproducibility of new generative models compared to non-generative ("predictive") AI models. This MI-CLAIM-GEN checklist also seeks to clarify cohort selection reporting with unstructured clinical data and adds additional items on alignment with ethical standards for clinical AI research.

**Part 1. Study design**
All elements of study design from the original MI-CLAIM checklist, including clear descriptions of the research question, cohort selection, and baseline values, remain applicable to all clinical AI studies. Here, we add clarifications to checklist items on study design choices, including reproducible cohort selection, for generative AI studies.

**Part 1A. Study design for generative modeling**
Use of generative models requires careful consideration of datasets, labels, evaluation, and interpretation of results. Similar to traditional machine learning, generative modeling tasks often involve prediction of either categorical or continuous outputs. In both cases, labels should follow previous supervised machine learning guidelines and be robust, clinically validated, and reflective of a clinical outcome of interest. How labels were derived should also be clearly documented, including the source of the labels and protocols to retrieve these labels. If labels are provided by human annotators, multiple annotators are suggested[7]. Details on annotation guidelines and inter-annotator agreement should be provided. For more complex, unstructured outputs, such as summaries of clinical notes, that do not readily map to simple labels, more robust evaluation frameworks are necessary, which may again involve both automated and human evaluation. We discuss these evaluation strategies in detail in part 4.

Researchers should also be careful of training data memorization ("data leakage" or "contamination")[8,9]. Almost all publicly available datasets are included in generative model training data and should not be used as test datasets unless it can be demonstrated that the model has not been trained on the specific task or the data was published after the model was trained[10]. Here, we take training to include any methods that update model weights, such as pretraining, finetuning, or reinforcement learning, as well as in-context learning, in which examples are presented to models as part of a prompt without changing model weights (Figure 1). One option to test for memorization is to see whether the generative model can regenerate large portions of the dataset[11]. Importantly, however, memorization of data is still possible even if the foundation model cannot regenerate the dataset in this way and should be listed as a limitation if public datasets are used for testing[12]. We additionally discuss the need for independent validation and test datasets in part 2 of this checklist.

**Part 1B. Best practices for cohort selection**
Robust cohort selection also remains crucial for all clinical AI studies. We provide checklist items to better emphasize reproducibility in cohort selection and discuss best practices for constructing cohorts using unstructured or multimodal data, which are increasingly being used in generative modeling studies. Ideally, code to select patient cohorts and raw individual-level data should be made available (particularly with new mandates from funding agencies, including the National Institutes of Health), but in cases where either is not possible, full details on the patient cohort selection should be provided, such as attrition charts. Ambiguous language, such as "patients diagnosed with diabetes were included," should be avoided in favor of more reproducible terms, such as "patients who had at least 2 of the following ICD-10 codes: E11.*, E13.*... were included". Publication of codelists should also be provided for transparency and replication.

If methods to select patients are based on the presence of certain values mentioned in clinical text, the list of keyword terms, regular expressions, or other selection criteria should be made available. If qualitative factors, such as manual chart review, are used to identify patient cohorts, these should be detailed and the qualifications (eg. years of practice, specialty, etc) of the reviewer should be reported. Pre- and post-processing steps, such as extracting specific sections, converting text to lowercase, lemmatization or stemming, and/or mapping to standard vocabularies, should also be reported in full.

If datasets are de-identified or are otherwise not representative of the clinical settings presented by the research question, these limitations should be described and discussed in detail. This information may include how dates were shifted to preserve privacy, whether age is masked, specific methods used for redaction of text, which Electronic Health Record (EHR) vendor the data was derived from, if the data were limited to specific department(s) and/or note types, or other limitations compared to real-world settings. Sensitivity analyses should be performed where appropriate to justify any patient selection criteria deviating from established guidelines, and whether the final cohort mirrors that of the real-world patient population (in terms of clinical characteristics, demographics, etc.). Specifications for handling missing data should also be provided, if applicable.

**Part 2. Data and resource assessment**
In addition to new model architectures, reporting on generative clinical models must also include additional information about external datasets or tools that a model may interact with through approaches like retrieval augmented generation[2] (RAG) or function calling[13]. We develop checklist items to reflect the inclusion of these different components of compound AI systems[14] (Figure 1), and again emphasize that all training, validation, and testing datasets are independent of each other or present any potential data leakage as a limitation.

Another difference to generative model research that we highlight here is in dataset preprocessing. While most traditional machine learning methods typically rely on large, well-annotated datasets for training, newer models have been shown to be capable of performing tasks with minimal examples (few-shot), or even without any specific examples (zero-shot). For supervised machine learning models, common data splits typically use about 70-80% of the data for training or model finetuning, about 10-20% for hyperparameter tuning, and the remainder used only for final model evaluation. In contrast, the training dataset can be kept to a minimal fraction of the data for few- or zero-shot approaches, although it should still be kept independent from the validation or test datasets. Data splits should be performed at the patient level, with all data from each patient only included in one of the splits to maintain independence.

If performing prompt engineering, which can be thought of as another hyperparameter to tune model performance, it is also important to tune the prompt on a "prompt validation" dataset that is kept separate from the final test dataset. Previous studies have used 5% of the data or a minimum of 50 to 100 samples[15,16] for these prompt validation datasets. While the same validation dataset should be used for prompt engineering between different models, the best prompt selected for each model may vary. As discussed in section 6 for end-to-end reproducibility, all prompts tested during prompt engineering should be shared verbatim should be reported, along with their performance and a discussion of robustness of the model relative to different prompts[17]. If models are deployed to interactive settings with user-provided prompts, these prompts and any resulting model variability should also be evaluated and discussed.

As prompt engineering is a rapidly evolving field, this checklist does not specify how to approach prompt development beyond the use of independent prompt validation datasets and appropriate randomization. We direct readers to follow best practice guidelines laid out by each model developer, which often emphasize using clear, descriptive, concise instructions, providing a value to output if the task is not applicable, and using leading cues to direct the formats of outputs[18–20]. For classification tasks where potential labels are provided in the prompt, the order of these labels should be randomly shuffled since models may be sensitive to the position of values in the prompt[21,22]. New approaches, such as chain-of-thought approaches for reasoning tasks[23], self-consistency with shuffling[24], or training vector representations as "soft prompts"[4], should be considered when developing prompts.

**Part 3. Baseline model selection**
Baseline model comparisons are important to provide controls for evaluating model performance. Generative model performance should be compared to rigorously selected baseline models, which may include other generative models but also non-generative approaches[15,25]. Given the rapid pace of model development, the most recent model available should be preferred for testing, while previous versions or methods can serve as baselines where appropriate. Performance, as well as the data, labeling, and computational resources required to train and test each model, should be reported in order to better measure each model's performance as well as efficiency[26,27].

For applications without previously developed methods, researchers should report performance relative to benchmarks set by naive models, such as a dummy classifier that predicts the majority class or a mean predictor for regression tasks. Other open source baselines are also strongly encouraged and researchers should consider evaluating models of different sizes if available.

Any post-processing of model and baseline outputs should be detailed in the methods, including how errors or unexpected outputs are handled. If large training datasets are used for baselines models compared to zero- or few-shot approaches for generative models, we encourage researchers to report their performance across various volumes of data. These training datasets, context lengths, and all other model details should be reported or clearly referenced to describe their potential impact on the task being tested. Discussion of the tradeoffs between compute and cost requirements is also encouraged. This allows an understanding of the scalability and efficiency of these non-generative models compared to their generative counterparts[15,27].

**Part 4. Evaluation of model performance**
Evaluation metrics for generative models should distinguish between metrics that measure *overlap accuracy*, which measures proportions of overlapping subunits (eg. tokens, pixels),

*semantic accuracy*, which compare the meanings of outputs and labels, and *clinical utility*, which measure how models affect clinical workflows or downstream patient outcomes[28–30]. We identify best practices for both automated and clinical expert evaluations, with a focus on metrics developed to handle the complex, unstructured outputs from generative models. We also emphasize the need for evaluation of models on real-world datasets that go beyond traditional benchmarking, which are often performed on curated datasets that are not reflective of real-world complexity and are also often publicly available and present in training datasets of many generative models.

If deployed to clinical settings, continuous evaluation and monitoring of these models is essential to identify any dataset shifts[31] or changes in model behavior, particularly if using black box systems where model versions may be modified without warning[32].

**Part 4A. Automated model evaluation**
Similar to traditional machine learning classification setups, accuracy, F1 scores (for imbalanced datasets), or other suitable metrics should be reported, along with class distribution, for categorical labels. For continuous outputs, such as time saved or changes to patient activity scores, which are also common for assessing *clinical utility* of models, best practice statistical approaches and reporting should be applied, including appropriate estimators for causal effects and multiple hypothesis testing.

For unstructured text outputs, automated *overlap scoring* methods like BLEU and ROUGE are commonly used, but these only capture how well tokens match between model predictions and a ground truth reference. These provide an estimate of how well the models produce text that look correct, but do not assess whether the answers are clinically accurate, so are often poorly correlated with human evaluation on biomedical tasks[33,34]. These methods also often fail in cases of negation[35], where the model produces values such as "correct" that can match a significant proportion of the negated value "not correct" but has the opposite meaning. Additionally, these methods may not be appropriate for certain clinical tasks where reference documents typically do not exist, such as in document summarization[36].

*Semantic scoring* methods, such as BERT-based scoring methods[37] or panels of similar metrics[38,39], are demonstrating initial promise on general, non-medical tasks [40–42]. However, rigorous evaluation is required before applying these approaches at scale on new, clinical tasks[36,43] and their credibility for the given study must be articulated if used. Some studies have also begun to use generative models for evaluation[42,44], but again, validation of these methods should be included and any limitations should be clearly stated in the discussion section.

**Part 4B. Human model evaluation**

Human model evaluation remains the gold-standard for assessing *semantic accuracy* and *clinical utility* of generative models. As much as possible, evaluation should be conducted in a blinded fashion, with Turing-like assessments against ground truth values or across multiple metrics to gauge the accuracy, appropriateness, bias, and other aspects of model performance[45,46]. For complex outputs or simulated scenarios, Objective Structured Clinical Examination (OSCE) type evaluations can be considered that assess model performance across multiple axes that better reflect real-world clinical encounters or workflows[33,47]. Although evaluations are dependent on the question being asked, we emphasize the need for multiple clinical reviewers and transparent reporting of inter-reviewer variability and formal evaluation guidelines used.

**Part 5. Examination of generative models**

**Part 5A: Interpretability and feature importance**
Interpretability research for generative models remains an active field of investigation, and we maintain suggestions from the original MI-CLAIM checklist to apply best-practice interpretability methods. These may include local interpretability techniques like LIME[48] and SHAP[49], gradient and attention analysis[50,51] for attributing importance scores to different input segments, probing methods to identify encoded knowledge[52], rule-based methods to explain model predictions as if-then-else rules[53], and counterfactual analysis to compare minimal example pairs for which language models exhibit different behavior[54].

Recently, methods like chain-of-thought have become popular for generating explanations of how a model might solve a problem to improve language model reasoning[23]. However, these generated explanations may not always align with model outputs and should not be used as a method of model interpretability[52,55]. Careful evaluation of these methods should be performed when applied to new clinical tasks[36,56], particularly since most of these methods were originally developed for models with shorter context lengths or less complex tasks.

Error analysis and sensitivity analysis (ablation tests), including prompt sensitivity tests, are also strongly encouraged as methods to better understand model behavior, particularly if evaluation datasets or models are not made publicly available. It is becoming increasingly important to understand how generative models may fail in clinical settings, which can provide insights into their capabilities and limitations beyond accuracy metrics. Continuous monitoring of model behavior, including interpretability, is again essential and researchers should include recommendations or discussion for post-deployment evaluation.

**Part 5B. Bias, privacy, and harm assessments**
Identifying potential harms of modeling approaches is also becoming increasingly important for generative models, which can produce complex, unstructured outputs that may be difficult to identify as inaccurate or biased[57,58]. The Generative MI-CLAIM checklist introduces new items that encourage discussion, identification, and mitigation of study biases, privacy concerns, and potential for harm. Here, we briefly discuss examples of approaches that may be used to promote transparency and inclusivity in these study design elements.

Models trained on biased data can perpetuate biases in generated content[59]. All available details regarding data distribution of any training and evaluation datasets should be reported, including patient sociodemographic information, any data imbalance, the time period when the data was collected, and any changes to best practice medical guidelines during this time period[60]. When possible, analysis of model performance across diverse patient subgroups and data subtypes[61] is strongly encouraged to identify biases in downstream deployment and impact on patient care and decision-making[8,9]. This is particularly critical if training or evaluation datasets are not reflective of real-world patient diversity or clinical workflows, and external validation to assess model fairness and robustness should be performed across different data distributions if possible. For assessment of cultural and social biases, researchers should consider engaging with a diverse set of clinical evaluators. Potential clinical impacts of generative models should also be identified or if possible, assessed in real-world settings with patient-centered approaches that are inclusive of diverse cultural and social communities[33,62].

Due to the rapid development of generative modeling approaches, data privacy and security vulnerabilities also remain of significant concern[5]. Models that may be deployed to real-world clinical care settings in particular must be evaluated for cybersecurity vulnerabilities, including adversarial prompt injections[63]. These vulnerabilities should be assessed based on up-to-date literature on privacy and security[64–66], and care must be taken to ensure that sensitive data or model outputs from sensitive data are maintained in secure environments[67]. This section provides only a brief description of potential approaches to analyzing and addressing model safety, fairness, and reliability, and we point researchers towards more comprehensive guidelines on each of these topics[64,68–70]. We also encourage the release of model weights, although these should be treated with the same care as clinical datasets, including the use of secure repositories and restricting access with model use agreements.

**Part 6. End-to-end pipeline replication**
Reproducible methods for generative modeling research should allow the community to replicate 1) data collection and cohort selection, 2) model development and inference, and 3) end-to-end evaluation. We add new checklist items to identify the level of transparency presented, with separate tiers for reproducible data processing and model training or usage.

For data and analytic transparency, all code and data should be provided in appropriate, accessible repositories. If full real-world datasets cannot be shared, a sample of the raw data, synthetic data, or the data structure derived following patient selection as well as the processed data should be provided[71]. Use of any synthetic data and strategies for generation should follow individual journal guidelines on data reporting. Along with datasets used and code used for analysis, we also emphasize the importance of releasing all prompts tested and corresponding results. Researchers should also include a requirements file, like a requirements.txt for Python packages, which lists all the dependencies and their precise versions, as different versions can produce different results. Additionally, the use of containerization tools like Docker can encapsulate the environment and further aid in replication efforts.

As part of model transparency, we add checklist items for researchers to include infrastructure and compute requirements needed to run or develop the model as part of their methods. These may include, but are not limited to, the type and quantity of hardware used, key dependencies, operating system, actual or estimated costs of inference or training, and training time if applicable. For reproducible model development or usage, any random seeds used and other hyperparameters should also be reported, along with detailed descriptions of model inputs, versions, and implementation frameworks, especially if code and/or data are not provided. External datasets, base model(s) used, embedding model(s), retrieval model(s), and other auxiliary models or tools, for example in retrieval augmented generation[3] or function calling[13] approaches, should also be disclosed (Figure 1), with discussion on which resources are static versus specific to the cohort or use case.

Drawing from best practices set out for all model development, the checklist also includes a section to report clinical model cards[68] or labels[72] that summarize the model capabilities, intended use, training data and limitations, potential biases, and model risks. An example of a clinical model card has been provided (Table 2). While the MI-CLAIM-GEN checklist summarizes whether the current clinical generative AI study has been conducted and reported using best-practice recommendations, model cards provide additional transparency around model development, intended uses, and known limitations to support the appropriate use of these models in future research or deployment.

**Conclusions**

There is enormous potential for generative models to unlock new research directions and applications, but robust study design and evaluations are crucial for developing reproducible, transparent, safe, and diverse models for clinical research and deployment. While the focus and examples here pertain primarily to generative language modeling, these principles can be applied to research using biomedical vision, speech, and multimodal models as well. This generative AI checklist begins to formalize guidelines for reporting on clinical generative modeling study design, baseline model development, evaluation, interpretability, and end-to-end reproducibility.

The MI-CLAIM-GEN checklist can be found on Github at the following link: https://github.com/BMiao10/MI-CLAIM-GEN. Since best practices for each aspect described are likely to change as new research emerges, the focus here is on the key differences in reporting for generative modeling compared to traditional AI model development.

We welcome continuous community feedback as the generative modeling landscape evolves, and also provide this space as a community forum for readers to identify and engage with best-practice approaches within each section of the MI-CLAIM-GEN checklist.

**Disclosures**
**BYM** is an employee at SandboxAQ. **IYC** is a minority shareholder in Apple, Amazon, Alphabet, and Microsoft. **SSu** is an employee at Ruby Robotics. **TZ** is a medical consultant for Xyla Health. **MG** is an employee of Pfizer, Inc. **AJB** is a co-founder and consultant to Personalis and NuMedii; consultant to Samsung, Mango Tree Corporation, and in the recent past, 10x Genomics, Helix, Pathway Genomics, and Verinata (Illumina); has served on paid advisory panels or boards for Geisinger Health, Regenstrief Institute, Gerson Lehman Group, AlphaSights, Covance, Novartis, Genentech, and Merck, and Roche; is a shareholder in Personalis and NuMedii; is a minor shareholder in Apple, Facebook, Alphabet (Google), Microsoft, Amazon, Snap, 10x Genomics, Illumina, CVS, Nuna Health, Assay Depot, Vet24seven, Regeneron, Sanofi, Royalty Pharma, AstraZeneca, Moderna, Biogen, Paraxel, and Sutro, and several other non-health related companies and mutual funds; and has received honoraria and travel reimbursement for invited talks from Johnson and Johnson, Roche, Genentech, Pfizer, Merck, Lilly, Takeda, Varian, Mars, Siemens, Optum, Abbott, Celgene, AstraZeneca, AbbVie, Westat, and many academic institutions, medical or disease specific foundations and associations, and health systems. Atul Butte receives royalty payments through Stanford University, for several patents and other disclosures licensed to NuMedii and Personalis. Atul Butte's research has been funded by NIH, Peraton (as the prime on an NIH contract), Genentech, Johnson and Johnson, FDA, Robert Wood Johnson Foundation, Leon Lowenstein Foundation, Intervalien Foundation, Priscilla Chan and Mark Zuckerberg, the Barbara and Gerson Bakar Foundation, and in the recent past, the March of Dimes, Juvenile Diabetes Research Foundation, California Governor's Office of Planning and Research, California Institute for Regenerative Medicine, L'Oreal, and Progenity. None of these organizations or companies had any influence or involvement in the development of this manuscript. **BN** is a co-founder at Qualified Health PBC. **All other authors** have no conflicts of interest to disclose.

**Table 1. MI-CLAIM-GEN checklist for generative AI clinical studies.**
Sections highlighted in orange indicate modifications to the original checklist and items highlighted in green indicate additions to the checklist described in this manuscript. Values that are not highlighted reflect original checklist items that continue to be applicable to all clinical AI studies.

| Before paper submission | | |
|---|---|---|
| **Study design (Part 1)** | **Page number** | **Notes if not completed** |
| The clinical problem in which the model will be employed is clearly detailed in the paper. | | |
| The research question is clearly stated. | | |
| All cohort selection criteria and study design are detailed in such a way that they can be reproduced by an external researcher. | | |
| Details on how labels were generated are described, including any annotation guidelines, level of experience of annotators, inter-annotator scores, etc. | | |
| Is the output data type categorical, continuous, or unstructured? | ☐ Categorical<br>☐ Continuous<br>☐ Unstructured | |
| The characteristics of the cohorts are detailed in the text and are shown to be representative of real-world clinical settings. | | |
| **Resources and optimization (Part 2)** | **Page number** | **Notes if not completed** |
| Model/application components are clearly detailed including: base model(s) used, embedding model(s), retrieval model(s), and other auxiliary models or tools. | | |
| The origin of all data sources for model training, finetuning, or inference is described and the original format is detailed in the paper. | | |
| All data preprocessing for model training, finetuning, or inference is described, including appropriate randomization and other transformations. | | |
| The independence between training, validation (including for prompt engineering), and test sets has been described, and data is split at the patient level. | | |
| **Model performance and evaluation (Parts 3-4)** | **Page number** | **Notes if not completed** |
| The state-of-the-art solution used as a baseline for comparison has been identified and detailed. | | |
| The performance comparison between the baseline and the proposed model is presented with the appropriate statistical significance. | | |
| Identify what evaluation(s) were performed, and provide clear justifications for the primary metrics used for each evaluation. | ☐ Overlap accuracy<br>☐ Semantic accuracy<br>☐ Clinical utility | |
| If applicable, details on human evaluation are described, including any evaluation guidelines, level of experience of evaluators, inter-reviewer scores, etc. | | |
| **Model examination (Part 5)** | **Page number** | **Notes if not completed** |
| Relevant interpretability techniques, error analysis, and/or other approaches are applied to demonstrate an absence of unreasonable risk and brittleness, including a low risk of catastrophic and especially undetected failure. | | |
| A discussion of the risk revealed by the examination results is presented with respect to model/algorithm performance. | | |
| Which step(s) have been taken to understand model biases, privacy and security concerns, and other potential harm? | ☐ Discussion<br>☐ Identification<br>☐ Mitigation | |
| A discussion and/or assessment of relevant distribution shifts and their impact on the model's performance has been provided | | |
| The authors provide recommendations or discussion of post-deployment evaluation | | |
| **Reproducibility (Part 6)** | **Page number** | **Notes** |
| **Data transparency: choose appropriate tier of transparency** | | |
| Tier 1: complete sharing of the code and data, including all prompts tested, software dependencies, and evaluation setups. | | |
| Tier 2A: complete sharing of the code with synthetic data provided | | |
| Tier 2B: complete sharing of the code | | |
| Tier 3: no sharing of code or data | | |
| **Model transparency** | | |
| Model hyperparameters, along with infrastructure and compute requirements for running and/or developing the model are included, specifying hardware type, costs, and training time where applicable. | | |
| A clinical model card is included summarizing the model capabilities, intended use, descriptions of any dataset or other integrations, limitations, potential biases, and risks | | |
| If applicable: Model weights are released to a secure repository and/or with appropriate use agreements. | | |

**Figure 1. Components of model training and inference to report for end to end replication.**
Independent datasets and data splits (validation, test) used during any stage of model training should all be reported. This includes any data used for in-context learning, such as databases used for retrieval augmented generation or any prompt engineering performed. Additionally, any post-processing, including external tool usage, should also be reported. Models merging multiple, existing models should provide components for each model.

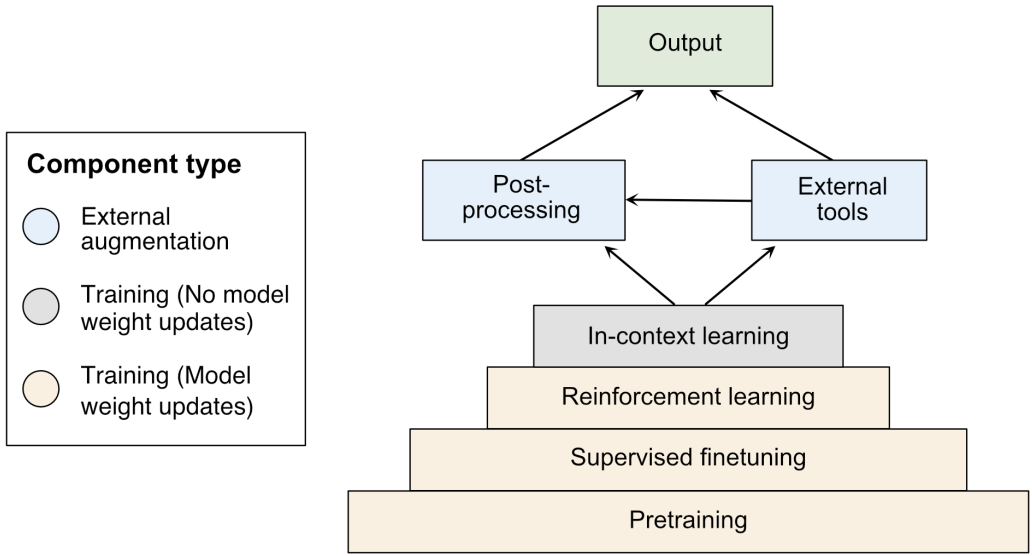

**Figure 2. Components of a clinical model card.**
An example model card, formatted as a clinical "model facts" label[72], for a fictional model created to assist in clinical decision support around sepsis diagnosis and management. The clinical model card should provide a summary of how a model was developed, intended use, out-of-scope uses, performance, limitations, and recommendations for safe deployment.

### Model Summary

**Model name:** Sepsis-GPT  **Developer:** MI-CLAIM Health
**FDA Clearance:** N/A  **Last updated:** June 20, 2024  **Version:** 1.0

### Intended usage
- **Indication for use:** Assisting emergency physicians in diagnosing and managing patients with suspected sepsis.
- **Out of scope uses (Contraindications):** Not intended for use in patients under 18 years old, pregnant women, or those with immunocompromised conditions

### Development
- **Pretrained model:** Clinical-T5 (Derived from T5-Base)
- **Pretraining dataset description:** Further pretrained on MIMIC-III and -IV clinical notes
- **Finetuning**
    - **Method:** Supervised finetuning using labeled clinical notes, vital signs, and laboratory data
    - **Dataset:** 50,000 emergency department visits with confirmed sepsis diagnoses and severity labels derived from electronic health records
    - **Target:** Binary sepsis diagnosis and multiclass severity assessment
- **Prompt engineering**
    - **Method:** Few-shot learning with 3 random examples of clinical notes and corresponding diagnoses/severity assessments
    - **Dataset:** Curated set of representative clinical note snippets from 100 patients with annotations.
    - **Target:** Accuracy of diagnosis and severity assessment, minimizing false negative.
- **External tools**
    - Sepsis-3 diagnostic criteria, SOFA score calculator, antibiotic recommendation engine

### Validation and performance

| Validation Type | AUC (Diagnosis) | F1 (Management) | Cohort size | Dataset | Citation |
|---|---|---|---|---|---|
| Internal (retrospective) | 0.83 | 0.56 | 1,000 | Link 1 | doi:#### |
| Internal (prospective) | NA | NA | NA | NA | NA |
| External (retrospective) | 0.68 | 0.63 | 2,500 | Link 2 | doi:#### |

- **Primary clinical metric:** Accuracy of sepsis diagnosis and appropriateness of management recommendations.
- **Continuous monitoring recommendations:** Weekly review of a random sample of 50 model outputs by a clinical expert to assess the quality, appropriateness, and bias

### Warnings
- **Risks Resulting from Bias Findings:** Potential underdiagnosis in patients from underrepresented racial/ethnic groups.
- **Risks Resulting from Clinical Findings:** False negative diagnoses could lead to delayed treatment; false positives could lead to overtreatment.
- **Other Known or Suspected Risks within the Intended Domain:** Model may underperform on cases with incomplete data or atypical presentations.

### Other information
- **Citation:** Placeholder et al. *Sepsis-GPT model for sepsis diagnosis using real-world clinical data. 2024.*
- **License:** MIT License